# Retractions Revisited: A Replication, Extension, and Mechanistic Evaluation of Linguistic Obfuscation Effects in Fraudulent Science

David M. Markowitz[1]

**Affiliations**:

[1] Associate Professor, Department of Communication, Michigan State University, East Lansing, MI USA (email: dmm@msu.edu)

## Abstract

Prior work suggests fraudulent scientists write their articles with higher rates of linguistic obfuscation than genuine scientists, with a possible mechanism suggesting obfuscation makes research more costly for a reader to analyze. This result has not undergone additional empirical scrutiny in the open science era, nor have other psychological mechanisms been evaluated. The current replication and extension addressed this opportunity by comparing 800 fraudulent and 800 genuine control papers (2013-2025). The results supported the linguistic obfuscation hypothesis, suggesting fraudulent papers were more obfuscated than genuine ones. A parallel mediation model evaluated three competing mechanisms, finding that reality monitoring, as measured through decreased rates of adjectives, best explained the link between fraudulent science and linguistic obfuscation. Cross-validated detection accuracy using obfuscation features (60.7%) and an elastic net of over 100 language dimensions (71.6%) were significantly greater than chance. Implications for scientific fraud and psychology of language research are discussed.

*Keywords*: scientific fraud, psychology of language, obfuscation, linguistic obfuscation, language

# Retractions Revisited: A Replication, Extension, and Mechanistic Evaluation of Linguistic Obfuscation Effects in Fraudulent Science

More than a decade after the replication crisis (Open Science Collaboration, 2015), scientific fraud remains a salient concern for authors, editors, and readers of academic research. Most estimates suggest scientific fraud — defined as the manipulation, fabrication, or falsification of data (LaFollette, 1992) — is relatively rare (Fanelli, 2009; Fang et al., 2012), but its presence in the academy has direct financial (Stern et al., 2014), reputational (Lu et al., 2013), and credibility costs (Abalkina et al., 2025). Even at relatively low base-rates (Fanelli, 2009; Fang et al., 2012), fraudulent research can distort the scientific record and undermine one's confidence in the knowledge-production process (Steen, 2011). Identifying ways to identify or mitigate academic deception remains an important challenge for scientific communities at large.

Several mitigation efforts have been proposed. For example, the open science movement (e.g., via efforts such as preregistration, open data, open materials) has offered a range of practical tools to reduce the probability of fraud in the literature (Munafò et al., 2017), while others advocate for changing the incentive structure of scientific publishing to avoid "publish or perish" dynamics that may motivate questionable research practices (Nosek et al., 2012). Remedies are worthwhile and deserve significant attention, though a more basic and fundamental question remains less understood in the academic literature on scientific fraud: How is scientific fraud communicated in research papers, and how do such patterns reflect the psychology of academic deceivers? The current study seeks to address this question directly in a replication and extension of prior work.

Specifically, the present paper is a replication and extension of scholarship that investigated the writing of nearly 250 fraudulent papers and 250 genuine papers in search of obfuscation signals (e.g., less readable writing, more abstraction). The evidence from Markowitz and Hancock (2016) suggested papers retracted for data fraud were more obfuscated than papers without data manipulation concerns and proposed that obfuscation occurs to make false claims more difficult to appraise. This work has provided an important foundation for understanding the verbal characteristics of scientific fraud, but it has not undergone additional empirical scrutiny or testing, especially after such an important period of academic history like the replication crisis. To this end, there are three important and interrelated research questions addressed in this work: (1) To what degree does obfuscation remain a signal of scientific fraud after the replication crisis?, (2) What are the potential mechanisms underlying obfuscation effects in fraudulent science?, and (3) How well can scientific fraud be detected, computationally? These questions are addressed in a replication and extension of the original work (Markowitz & Hancock, 2016), using four times the number of cases and exploring two additional mechanisms.

**Examining Scientific Fraud: Support from the Linguistic Obfuscation Hypothesis**

The idea of investigating how fraudulent science modifies the writing of academic papers is theoretically rooted in psychology of language scholarship, which treats words as markers of attention (Boyd & Markowitz, 2025; Boyd & Schwartz, 2021; Pennebaker, 2011). For example, people who communicate with high rates of negative emotion terms (e.g., *hate*, *disgust*) are attending to a negative emotional state compared to people with low rates of negative emotion terms; people who use high rates of self-references (e.g., *I*, *myself*) are attending more to the self than those who use low rates of self-references. This *words-as-attention* model of language use has had massive appeal and utility in the social sciences (Boyd & Markowitz, 2025; Boyd & Schwartz, 2021; Tausczik & Pennebaker, 2010), with evidence suggesting words connect to a range of social and psychological dynamics including personality (Ireland & Mehl, 2014; Kern et al., 2014), social status (Kacewicz et al., 2014), depression (Tackman et al., 2019), intergroup bias (Maass, 1999), and many others, including the topic of the present work, deception (e.g., Hauch et al., 2015; Markowitz & Hancock, 2016; Toma & Hancock, 2012).

What language dimensions betray scientific fraud? Recall, one study investigating this phenomenon (Markowitz & Hancock, 2016) drew on the linguistic obfuscation hypothesis from the financial fraud literature to identify candidate metrics (Bloomfield, 2002; Courtis, 1998; Humpherys et al., 2011; Li, 2008). The linguistic obfuscation hypothesis suggests companies with evidence to hide often write their end-of-year reports in a more obfuscated manner than companies without evidence to hide, and there are five language parameters associated with linguistic obfuscation. First, fraudulent reports are often less readable than genuine reports in an effort to make such reports more difficult to understand and parse (Li, 2008). Longer words and sentences are often more difficult for people to understand and appraise (Shulman et al., 2024), which might help deceivers avoid detection. Second, fraudulent reports are often written more abstractly than genuine reports — containing lower rates of articles (e.g., *a*, *the*), prepositions (e.g., *above*, *below*), and quantifiers (e.g., *more*, *less*) (Larrimore et al., 2011) — which reflects an interest in making deceptive speech more vague and difficult to scrutinize (Sun & Papadokonstantaki, 2023). Third, fraudulent reports often contain a lower rate of positive emotion terms compared to genuine reports because authors are typically aware that their results are “too good to be true,” and deceivers often have difficulty approximating the appropriate level of genre-specific language characteristics like emotion in their language (Hauch et al., 2015; Newman et al., 2003; Ott et al., 2011). Fourth, fraudulent reports often contain more jargon than genuine reports as authors try to appear more credible, technical, and scientific in papers that have little evidentiary basis (Berrios et al., 2025; Markowitz & Hancock, 2014). Finally, in a related manner, fraudulent reports often contain a higher rate of causal terms (e.g., *because*, *affect*), which may reflect fraudulent scientists' need to construct elaborate explanations for claims that lack such an evidentiary basis compared to genuine scientists.

Indeed, Markowitz and Hancock (2016) observed that the writing of fraudulent science is characterized by higher rates of abstraction, causal terms, and jargon, but lower rates of readability and positive emotion terms. Subsequent work suggests linguistic obfuscation can have consequences beyond signaling scientific misconduct. Less readable and more complex disclosures can alter how people process and respond to information, including weakening or delaying investor reactions in corporate settings (Rennekamp, 2012; You & Zhang, 2009), making information more difficult for audiences to extract and evaluate (Bushee et al., 2018), and influencing individual investors' investment choices and their subsequent investment performance (Lawrence, 2013). Obfuscation can also have interpersonal and ethical consequences. That is, companies using more (versus less) obfuscated values statements (e.g., codes of conduct) were perceived as less warm and moral, and exposure to highly obfuscated corporate values statements can increase readers' own deceptive behavior in the form of cheating on math and word tests (Markowitz et al., 2021). Together, these findings suggest linguistic obfuscation is a critical marker of deception, in general, and it can meaningfully alter how people evaluate, understand, and respond to the deceiver.

**The Current Paper**

Against this backdrop, the current paper seeks to replicate and extend prior work that examined how fraudulent papers are written relative to genuine papers. Replications of this type are needed because it is unclear how well these results hold over time (especially after the replication crisis), and establishing the credibility of extant evidence is essential for the utility of our social sciences (Levine & Markowitz, 2024). This work is extended in two ways as well. First, possible mechanisms for the relationship between scientific fraud and linguistic obfuscation are examined beyond those established in the literature (e.g., attempting to appear more scientific by increasing the number of references per paper, which make the science more difficult to appraise; Markowitz & Hancock, 2016). Two new mechanisms — psychological distancing and reality monitoring — are examined alongside academic references to assess why scientific fraud might be linked with linguistic obfuscation. Finally, this paper used several techniques to evaluate how well fraudulent and genuine papers can be classified. A top-down, theory-driven approach using the obfuscation dimensions and a bottom-up, exploratory approach using over 100 verbal features complemented each other to determine how well scientific fraud could be discriminated using language dimensions alone.

## Method

### Data Collection

To evaluate the linguistic difference between fraudulent science and genuine science controls from 2013-2025, data from three sources were combined to create an initial list of retractions. The Retraction Watch Database (2026) catalogs retractions across the sciences and includes a range of metadata (e.g., the reason for retraction, publication and retraction dates, authorship information). From the 67,264 records available as of September 2026, articles were retained if they contained an original publication year of 2013 or later and with a PubMed identifier, yielding an initial list of 23,647 retractions. Each record was then merged with PubMed metadata for both the original paper and its retraction notice. Two additional sources were used to capture retractions absent from the Retraction Watch Database. First, PubMed was queried for retracted publications, which identified 6,045 additional records. Second, the CrossRef API was queried for retractions communicated directly by publishers, which yielded an additional 20,044 records. Data were merged and deduplicated across sources using each paper's DOI or PubMed identifier. The final corpus contained 43,261 retracted papers, though only 16,663 of them had retraction notices and were coded for the presence or absence of fraud.

**Fraud Identification**

Two large language models (LLMs: *gpt-4o* and *claude-sonnet-5*) coded each of the 16,663 retraction notices into one of three categories: (1) fraud (e.g., data fabrication, data falsification, image manipulation, or a finding of misconduct), (2) not fraud (e.g., honest error, plagiarism, ethical or authorship issues, publisher error), or (3) unclear (e.g., a determination could not be made). The exact LLM prompt is in the online supplement. Agreement between the two models was almost perfect (Cohen's $\kappa = 0.945$). Disagreements ($n = 297$) were resolved by a third model (*claude-opus-5-5*). This process identified 1,540 papers retracted for fraud.

Consistent with prior work (Markowitz & Hancock, 2016), each fraudulent paper was yoked to one unretracted and genuine control paper from PubMed matched on journal and publication year, plus 98% of pairs were also matched on topic (via MeSH descriptors, or standardized terms used to categorize biomedical articles). Candidates were excluded if they appeared anywhere in the full retraction corpus or had already been assigned as a control, and the control was selected at random from the list of eligible papers. Full texts were retrieved with main text separated from reference lists. The final corpus included 800 pairs ($n = 800$ fraudulent vs. $n = 800$ genuine controls) whose full texts could be obtained.

**Automated Text Analysis**

All texts were analyzed with Linguistic Inquiry and Word Count (LIWC) (Pennebaker et al., 2022), an automated text analysis program that counts words as the percentage of the total word count across many dimensions in its internal dictionary. Five dimensions comprising the linguistic obfuscation index were measured, including positive loadings for (1) jargon (words outside of the LIWC dictionary, or 100-Dictionary), (2) abstraction, or the standardized inverse sum of articles, prepositions, and quantifiers, and (3) causal terms like *because* and *depend*; negative loadings were calculated in the obfuscation index for (4) positive tone words like *admire* and *intimate*, and (5) readability using the Flesch Reading Ease metric (Benoit et al., 2021; Flesch, 1948). Each dimension was standardized prior to the obfuscation index calculation.

# Results

Descriptive statistics and intercorrelations between linguistic obfuscation variables are in Table 1. The results suggest, consistent with prior work (Markowitz & Hancock, 2016), papers retracted for fraud were written with a greater rate of linguistic obfuscation than genuine controls (Cohen's $d = 0.27$). Results were statistically significant and in the predicted direction for jargon (Cohen's $d = 0.23$), abstraction (Cohen's $d = 0.28$), causal terms (Cohen's $d = 0.16$), and positive tone (Cohen's $d = 0.28$)[1], but fraudulent papers were unexpectedly more readable than genuine controls (Cohen's $d = 0.35$). These results were also largely consistent over time (see Figure 1).

## Psychological Mechanisms

Consistent with the idea that fraudulent scientists may attempt to make their work more difficult to appraise than genuine scientists (Humpherys et al., 2011), obfuscation was positively associated with the natural log-transformed number of references in each paper ($r = .134$, $p < .001$). How well does this result explain the obfuscation effect relative to other potential mechanisms? Two additional mechanisms were tested. The first is psychological distancing (Newman et al., 2003), which argues that for some people, lying can be stressful and the potential consequences of being caught in a deception may motivate people to psychologically "move away" from the deceptive event (e.g., which is often associated with a reduction in self-references in false versus truthful speech). Psychological distancing was measured according to established work (Cohn et al., 2004).[2] The second mechanism concerns reality monitoring (Johnson & Raye, 1981), which suggests false experiences often contain fewer details than genuine experiences because false experiences lack evidentiary grounding in actual events. Reality monitoring was measured through rates of adjectives (Markowitz & Hancock, 2014).

[1] Results were consistent if positive emotion terms (Cohen's $d = 0.36$) were used instead of positive tone.

A parallel mediation model, predicting linguistic obfuscation from article type (i.e., fraud vs. genuine control) and three parallel mediators (i.e., references, psychological distancing, adjectives), was estimated with standard errors clustered on matched pairs and 95% Monte Carlo confidence intervals (20,000 draws) for the indirect effects. The mediators and obfuscation were standardized. The total effect of article type on obfuscation was significant ($B = 0.27$, $SE = 0.05$, $z = 5.78$, $p < .001$, 95% CI [0.18, 0.36]). Adjectives were the only substantial mediator, as fraudulent papers contained fewer adjectives than genuine papers ($B = -0.53$, $SE = 0.04$, $z = -12.19$, $p < .001$), and fewer adjectives were associated with more obfuscation ($B = -0.45$, $SE = 0.03$, $z = -14.80$, $p < .001$), producing a reliable indirect effect ($B = 0.24$, 95% CI [0.19, 0.29]) that accounted for 86% of the total effect. The indirect effect through psychological distancing was also reliable but negligible ($B = 0.01$, 95% CI [0.002, 0.020]), where fraudulent papers had a greater rate of psychological distancing than genuine papers ($B = 0.16$, $SE = 0.04$, $z = 3.84$, $p < .001$), though psychological distancing was only weakly related to obfuscation ($B = 0.06$, $SE = 0.02$, $z = 2.76$, $p = .006$). The indirect effect through references, which reflects the proposed mechanism by Markowitz and Hancock (2016), was reliable but negligible in size and opposite in sign relative to the other mediators ($B$ = -0.009, 95% CI [-0.019, -0.001]). Fraudulent papers contained fewer references than genuine papers ($B$ = -0.10, $SE$ = 0.05, $z$ = -2.27, $p$ = .023), and references were positively associated with obfuscation ($B$ = 0.09, $SE$ = 0.03, $z$ = 3.27, $p$ = .001). The indirect effect through adjectives was larger than the indirect effects through psychological distancing (*diff* = 0.22, 95% CI [0.18, 0.28]) and references (*diff* = 0.24, 95% CI [0.20, 0.29]), and the direct effect of article type on obfuscation was no longer significant with the mediators in the model ($B$ = 0.04, $SE$ = 0.04, $z$ = 0.83, $p$ = .407). Together, these data suggest obfuscation in fraudulent science is best explained by reality monitoring and a reduction in details.

[2] The psychological distancing formula is: z(article) + z(BigWords) – z(i) – z(we) – z(discrep) – z(focuspresent). Both self-references and collective references were used because authors might distance the self as either an individual or as an authorship group depending on the number of authors on a paper.

**Classification**

How well can the obfuscation features classify fraudulent versus control papers? Using a 10-fold cross-validated model, the five obfuscation features achieved 60.7% detection accuracy, an effect greater than chance ($p$ < .001, Cohen's $h$ = 0.216).[3] To benchmark this accuracy against other possible language dimensions that might separate fraudulent from genuine science, an elastic net logistic regression (mixing parameter $\alpha$ = .50) was constructed using all LIWC dimensions except for word count and punctuation terms. The penalty parameter was tuned through nested 10-fold cross-validation within each training set, so that accuracy estimates reflect only held-out texts.

This model achieved 71.6% detection accuracy (sensitivity, or true-positive rate = 75.1%, specificity, or true-negative rate = 68.0%, AUC = .784), a rate also greater than chance ($p$ < .001, Cohen's $h$ = 0.447). The five obfuscation features therefore accounted for approximately half of the above-chance accuracy obtained from the full LIWC feature set.[4] The strongest predictors in the full model were fewer differentiation words (e.g., *but*, *without*), more power words (e.g., *own*, *allow*), a greater focus on the past (e.g., *was*, *been*), and fewer adjectives (e.g., *more*, *new*) in fraudulent papers compared to genuine papers.

[3] Folds were assigned by matched pair so that a fraudulent paper and its control were never split across training and test sets.

An open vocabulary model (75.6%), which derives its features from the unigrams and bigrams in the texts rather than from prescribed categories, only improved on the full dictionary-based LIWC model by about three percentage points (McNemar's $\chi^2 = 8.56$, $p = .003$). A document embedding model (73.9%), which represented each paper as a 768-dimensional vector from a language model pretrained on scientific papers (SPECTER) rather than as counts of words or categories (Cohan et al., 2020), did not differ from the full LIWC model ($\chi^2 = 1.79$, $p = .180$; see Table 2). However, Area Under the Curve (AUC) metrics demonstrated that the open vocabulary model was superior to other statistical classification approaches.

## Open Science Characteristics of Fraudulent Papers

Finally, in an exploratory analysis, the full texts of each article were evaluated using the *ODDPub* package in *R* to identify open research practices in biomedical science articles (Riedel et al., 2020). This package identifies, in a binary manner (i.e., presence vs. absence), if a paper contains open science practices like open data and open code.

As expected, using a binary logistic regression model, the results suggested papers retracted for fraud had a lower probability of containing open data than genuine papers ($B = -1.00$, $SE = 0.39$, $z = -2.53$, $p = .011$). There was no statistically significant relationship between article type (fraud vs. genuine) and open code ($z = -0.30$, $p = .763$).

# Discussion

[4] This is calculated in the following manner: 60.7% - 50% = 10.7% above chance; 71.6% - 50% = 21.6% above chance; 10.7 / 21.6 = 0.50.

The current paper aimed to replicate and extend an original finding that suggested fraudulent science is written in a more obfuscated manner than genuine science, with a possible mechanism suggesting the reason for this effect is to make the science more difficult to scrutinize for readers (Markowitz & Hancock, 2016). The present study had nearly four times the number of cases than the original investigation and explored two additional mechanisms, with results replicating the original work and finding a more robust pathway to describe why fraudulent science may be communicated. That is, consistent with research on reality monitoring (Johnson & Raye, 1981), false experiences have little evidentiary grounding and therefore are described in less detail than genuine experiences, and fraudulent science was written in less detail and with fewer adjectives than genuine science. This work therefore offers important evidence to demonstrate the reliability of the original Markowitz and Hancock (2016) result and offers an important theoretical extension by describing new reasons why fraudulent science may be communicated in an obfuscated manner. Deception detection accuracy was also significantly greater than chance for the five obfuscation features alone and for the full LIWC dictionary.

There are several contributions of this work that are worth highlighting. First, this paper provides evidence that the linguistic obfuscation effect is reliable. With a larger and more recent sample of papers than the original investigation (Markowitz & Hancock, 2016), fraudulent science was again written in a more obfuscated manner than genuine science, and this pattern was largely consistent over time in the sample. In fact, four of the five obfuscation dimensions replicated in the predicted direction. Replications of this type are important because they establish which findings in the deception and language literature deserve continued theoretical attention. Not all effects replicated, however, including the result that fraudulent papers were more readable than genuine papers. One possibility for this result is that the nature of scientific fraud has changed over time. For example, papers produced by paper mills (Abalkina et al., 2025) may be written in a formulaic and simple manner that is easy to read, yet still vague and lacking in detail (e.g., prior work suggests scientific writing generated by non-humans like LLMs tend to be simpler than human scientific writing; Markowitz, 2024). This explanation is speculative and deserves more empirical attention, but it suggests readability may be a less stable marker of scientific fraud than the other dimensions.

Second, the current work offers a theoretical extension by evaluating *why* fraudulent science is linked to obfuscation. The original study proposed that fraudulent scientists obfuscate to make their papers more costly to analyze, with references serving as one way to make false claims more difficult to appraise (Markowitz & Hancock, 2016). The evidence did not support this account as the sole or primary mechanism. While references were positively associated with obfuscation, the link between scientific fraud and obfuscation was best accounted for by reality monitoring (Johnson & Raye, 1981). That is, fraudulent papers contained fewer adjectives than genuine papers, and papers with fewer adjectives were more obfuscated. This pattern suggests obfuscation in fraudulent science may be less of a deliberate attempt to conceal and more of an outcome of describing data, procedures, and results that have little or no grounding. People who report on events that did not occur have fewer details to draw upon (Johnson & Raye, 1981), and the same appears to be true for scientists who report on research that did not occur as described (Markowitz & Hancock, 2014). Together, these findings suggest a shift in how linguistic obfuscation should be understood theoretically as it may reflect what fraudsters lack (e.g., details) more than what they intend to hide.

Third, this paper offers a benchmark for how well scientific fraud can be detected from language alone. The five obfuscation features classified fraudulent and genuine papers at a rate greater than chance (60.7%), and a model with the full LIWC dictionary performed better (71.6%). The theory-driven features, therefore, provide a parsimonious and interpretable foundation for detecting scientific fraud. At the same time, the strongest predictors in the full model came from outside of the obfuscation hypothesis, which suggests there are other verbal dimensions of scientific fraud that current theory does not yet explain. It is important to note that these accuracies are not high enough to forensically detect future fraudulent science. Language patterns are *probabilistic signals* of fraud that can complement, but not replace, the investigative work of editors, institutions, and integrity experts to identify scientific fraud.

Finally, an exploratory analysis connected the language of scientific fraud to open science practices. Papers retracted for fraud were less likely to contain open data than genuine papers, a result consistent with the idea that transparency and fraud are in conflict with each other (Munafò et al., 2017). Authors with fabricated or manipulated data have more to lose by sharing fake evidence. The absence of open data is also conceptually consistent with reality monitoring, as fraudulent papers appear to provide less detail in their writing and less evidence in their materials. Together, these data suggest the open science movement and research on the language of scientific fraud are complementary efforts, with each offering signals about the credibility of published research.

**Limitations and Future Directions**

There are several limitations of this paper worthy of discussion. First, the papers in this corpus are primarily biomedical science articles. It is unclear how papers from other fields or disciplines might reveal scientific fraud, but this is a worthwhile consideration in future work. Second, the results in this work are not direct cause and effect, and the effect sizes are small but consistent with prior deception scholarship (Hauch et al., 2015). Third, while the classification analyses suggest the theory-driven and exploratory models improve above chance, they are likely ineffective for prospective deception detection. More cases and even more sophisticated word-counting approaches would be necessary to improve classification accuracy in this regard.

Future work might also benefit from analyses that measure how different forms of fraud modify the results, in addition to within-author study designs to measure how for the same author or authorship group accused of fraud, how the results manifest. It is also important to consider how, with the rise of generative artificial intelligence, new forms of academic misconduct or questionable research practices (e.g., having an LLM write full sections of a paper) might appear in the literature and modify how science is reported on over time.

**References**

Abalkina, A., Aquarius, R., Bik, E., Bimler, D., Bishop, D., Byrne, J., Cabanac, G., Day, A., Labbé, C., & Wise, N. (2025). 'Stamp out paper mills'—Science sleuths on how to fight fake research. *Nature*, *637*(8048), 1047–1050. https://doi.org/10.1038/d41586-025-00212-1

Benoit, K., Watanabe, K., Wang, H., Lua, J. W., Kuha, J., & European Research Council. (2021). *quanteda.textstats: Textual statistics for the quantitative analysis of textual data* (Version 0.95) [Computer software]. https://CRAN.R-project.org/package=quanteda.textstats

Benoit, K., Watanabe, K., Wang, H., Nulty, P., Obeng, A., Müller, S., & Matsuo, A. (2018). quanteda: An R package for the quantitative analysis of textual data. *Journal of Open Source Software*, *3*(30), 774. https://doi.org/10.21105/joss.00774

Berrios, T., Hu, D.-Y., & Vaid, J. (2025). The power of technical language: Does jargon use influence the credibility of misinformation? *Applied Cognitive Psychology*, *39*(6), e70137. https://doi.org/10.1002/acp.70137

Bloomfield, R. J. (2002). The "incomplete revelation hypothesis" and financial reporting. *Accounting Horizons*, *16*(3), 233–243.

Boyd, R. L., & Markowitz, D. M. (2025). Verbal behavior and the future of social science. *American Psychologist*, *80*(3), 411–433. https://doi.org/10.1037/amp0001319

Boyd, R. L., & Schwartz, H. A. (2021). Natural language analysis and the psychology of verbal behavior: The past, present, and future states of the field. *Journal of Language and Social Psychology*, *40*(1), 21–41. https://doi.org/10.1177/0261927X20967028

Bushee, B. J., Gow, I. D., & Taylor, D. J. (2018). Linguistic complexity in firm disclosures: Obfuscation or information? *Journal of Accounting Research*, *56*(1), 85–121. https://doi.org/10.1111/1475-679X.12179

Cohan, A., Feldman, S., Beltagy, I., Downey, D., & Weld, D. (2020). SPECTER: Document-level representation learning using citation-informed transformers. In D. Jurafsky, J. Chai, N. Schluter, & J. Tetreault (Eds.), *Proceedings of the 58th Annual Meeting of the Association for Computational Linguistics* (pp. 2270–2282). Association for Computational Linguistics. https://doi.org/10.18653/v1/2020.acl-main.207

Cohn, M. A., Mehl, M. R., & Pennebaker, J. W. (2004). Linguistic markers of psychological change surrounding September 11, 2001. *Psychological Science*, *15*(10), 687–693. https://doi.org/10.2307/40064028

Courtis, J. K. (1998). Annual report readability variability: Tests of the obfuscation hypothesis. *Accounting, Auditing & Accountability Journal*, *11*(4), 459–459.

Fanelli, D. (2009). How many scientists fabricate and falsify research? A systematic review and meta-analysis of survey data. *PLOS ONE*, *4*(5), 1–11.

Fang, F. C., Steen, R. G., & Casadevall, A. (2012). Misconduct accounts for the majority of retracted scientific publications. *Proceedings of the National Academy of Sciences of the United States of America*, *109*(42), 17028–17033.

Flesch, R. (1948). A new readability yardstick. *Journal of Applied Psychology*, *32*(3), 221–233. https://doi.org/10.1037/h0057532

Hauch, V., Blandón-Gitlin, I., Masip, J., & Sporer, S. L. (2015). Are computers effective lie detectors? A meta-analysis of linguistic cues to deception. *Personality and Social Psychology Review*, *19*(4), 307–342. https://doi.org/10.1177/1088868314556539

Humpherys, S. L., Moffitt, K. C., Burns, M. B., Burgoon, J. K., & Felix, W. F. (2011). Identification of fraudulent financial statements using linguistic credibility analysis. *Decision Support Systems*, *50*(3), 585–594. https://doi.org/10.1016/j.dss.2010.08.009

Ireland, M. E., & Mehl, M. (2014). Natural language use as a marker of personality. In T. M. Holtgraves (Ed.), *The oxford handbook of language and social psychology* (pp. 201–218). Oxford University Press.

Johnson, M. K., & Raye, C. L. (1981). Reality monitoring. *Psychological Review*, *88*(1), 67–85. https://doi.org/10.1037/0033-295X.88.1.67

Kacewicz, E., Pennebaker, J. W., Davis, M., Jeon, M., & Graesser, A. C. (2014). Pronoun use reflects standings in social hierarchies. *Journal of Language and Social Psychology*, *33*(2), 125–143. https://doi.org/10.1177/0261927x13502654

Kern, M. L., Eichstaedt, J. C., Schwartz, H. A., Dziurzynski, L., Ungar, L. H., Stillwell, D. J., Kosinski, M., Ramones, S. M., & Seligman, M. E. P. (2014). The online social self: An open vocabulary approach to personality. *Assessment*, *21*(2), 158–169. https://doi.org/10.1177/1073191113514104

Kjell, O., Giorgi, S., & Schwartz, H. A. (2023). The text-package: An R-package for analyzing and visualizing human language using natural language processing and transformers. *Psychological Methods*, *28*(6), 1478–1498. https://doi.org/10.1037/met0000542

LaFollette, M. C. (1992). *Stealing into print: Fraud, plagiarism, and misconduct in scientific publishing*. University of California Press.

Larrimore, L., Jiang, L., Larrimore, J., Markowitz, D. M., & Gorski, S. (2011). Peer to peer lending: The relationship between language features, trustworthiness, and persuasion success. *Journal of Applied Communication Research*, *39*(1), 19–37. https://doi.org/10.1080/00909882.2010.536844

Lawrence, A. (2013). Individual investors and financial disclosure. *Journal of Accounting and Economics*, *56*(1), 130–147.

Levine, T. R., & Markowitz, D. M. (2024). The role of theory in researching and understanding human communication. *Human Communication Research*, *50*(2), 154–161. https://doi.org/10.1093/hcr/hqad037

Li, F. (2008). Annual report readability, current earnings, and earnings persistence. *Journal of Accounting and Economics*, *45*(2–3), 221–247. https://doi.org/10.1016/j.jacceco.2008.02.003

Lu, S. F., Jin, G. Z., Uzzi, B., & Jones, B. (2013). The retraction penalty: Evidence from the web of science. *Scientific Reports*, *3*, 1–4.

Maass, A. (1999). Linguistic intergroup bias: Stereotype perpetuation through language. *Advances in Experimental Social Psychology*, *31*, 79–121. https://doi.org/10.1016/S0065-2601(08)60272-5

Markowitz, D. M. (2024). From complexity to clarity: How AI enhances perceptions of scientists and the public's understanding of science. *PNAS Nexus*, *3*(9), pgae387. https://doi.org/10.1093/pnasnexus/pgae387

Markowitz, D. M., & Hancock, J. T. (2014). Linguistic traces of a scientific fraud: The case of Diederik Stapel. *PLOS ONE*, *9*(8), e105937–e105937. https://doi.org/10.1371/journal.pone.0105937

Markowitz, D. M., & Hancock, J. T. (2016). Linguistic obfuscation in fraudulent science. *Journal of Language and Social Psychology*, *35*(4), 435–445. https://doi.org/10.1177/0261927X15614605

Markowitz, D. M., Kouchaki, M., Hancock, J. T., & Gino, F. (2021). The deception spiral: Corporate obfuscation leads to perceptions of immorality and cheating behavior. *Journal of Language and Social Psychology*, *40*(2), 277–296. https://doi.org/10.1177/0261927X20949594

Munafò, M. R., Nosek, B. A., Bishop, D. V. M., Button, K. S., Chambers, C. D., Percie du Sert, N., Simonsohn, U., Wagenmakers, E.-J., Ware, J. J., & Ioannidis, J. P. A. (2017). A manifesto for reproducible science. *Nature: Human Behaviour*, *1*, 1–9. https://doi.org/10.1038/s41562-016-0021

Newman, M. L., Pennebaker, J. W., Berry, D. S., & Richards, J. M. (2003). Lying words: Predicting deception from linguistic styles. *Personality and Social Psychology Bulletin*, *29*(5), 665–675. https://doi.org/10.1177/0146167203029005010

Nosek, B. A., Spies, J. R., & Motyl, M. (2012). Scientific utopia: II. Restructuring incentives and practices to promote truth over publishability. *Perspectives on Psychological Science*, *7*(6), 615–631. https://doi.org/10.1177/1745691612459058

Open Science Collaboration. (2015). Estimating the reproducibility of psychological science. *Science*, *349*(6251). https://doi.org/10.1126/science.aac4716

Ott, M., Choi, Y., Cardie, C., & Hancock, J. T. (2011). Finding deceptive opinion spam by any stretch of the imagination. *Proceedings of the 49th Annual Meeting of the Association for Computational Linguistics: Human Language Technologies*, *1*, 309–319.

Pennebaker, J. W. (2011). *The secret life of pronouns: What our words say about us*. Bloomsbury Press.

Pennebaker, J. W., Boyd, R. L., Booth, R. J., Ashokkumar, A., & Francis, M. E. (2022). *Linguistic inquiry and word count: LIWC-22* [Computer software]. Pennebaker Conglomerates. https://www.liwc.app

Rennekamp, K. (2012). Processing fluency and investors' reactions to disclosure readability. *Journal of Accounting Research*, *50*(5), 1319–1354. https://doi.org/10.1111/j.1475-679X.2012.00460.x

Retraction Watch Database. (2026). *Retraction Watch Database*. https://retractiondatabase.org/RetractionSearch.aspx?

Riedel, N., Kip, M., & Bobrov, E. (2020). ODDPub – a text-mining algorithm to detect data sharing in biomedical publications. *Data Science Journal*, *19*(1). https://doi.org/10.5334/dsj-2020-042

Shulman, H. C., Markowitz, D. M., & Rogers, T. (2024). Reading dies in complexity: Online news consumers prefer simple writing. *Science Advances*, *10*(23), adn2555. https://doi.org/10.1126/sciadv.adn2555

Steen, R. G. (2011). Retractions in the scientific literature: Is the incidence of research fraud increasing? *Journal of Medical Ethics*, *37*(4), 249–253. https://doi.org/10.1136/jme.2010.040923

Stern, A. M., Casadevall, A., Steen, R. G., & Fang, F. C. (2014). Financial costs and personal consequences of research misconduct resulting in retracted publications. *eLife*, *3*, e02956. https://doi.org/10.7554/eLife.02956

Sun, K.-K., & Papadokonstantaki, S. (2023). Lying aversion and vague communication: An experimental study. *European Economic Review*, *160*, 104611. https://doi.org/10.1016/j.euroecorev.2023.104611

Tackman, A. M., Sbarra, D. A., Carey, A. L., Donnellan, M. B., Horn, A. B., Holtzman, N. S., Edwards, T. S., Pennebaker, J. W., & Mehl, M. R. (2019). Depression, negative emotionality, and self-referential language: A multi-lab, multi-measure, and multi-language-task research synthesis. *Journal of Personality and Social Psychology*, No Pagination Specified-No Pagination Specified. https://doi.org/10.1037/pspp0000187

Tausczik, Y. R., & Pennebaker, J. W. (2010). The psychological meaning of words: LIWC and computerized text analysis methods. *Journal of Language and Social Psychology*, *29*(1), 24–54. https://doi.org/10.1177/0261927X09351676

Toma, C. L., & Hancock, J. T. (2012). What lies beneath: The linguistic traces of deception in online dating profiles. *Journal of Communication*, *62*(1), 78–97. https://doi.org/10.1111/j.1460-2466.2011.01619.x

You, H., & Zhang, X. (2009). Financial reporting complexity and investor underreaction to 10-K information. *Review of Accounting Studies*, *14*(4), 559–586. https://doi.org/10.1007/s11142-008-9083-2

**Table 1**

*Descriptive Statistics and Intercorrelations Between Obfuscation Dimensions*

| Variable | *M* | *SD* | 1 | 2 | 3 | 4 |
|---|---|---|---|---|---|---|
| 1. Abstraction | 0.00 | 1.91 | -- | | | |
| 2. Jargon (%) | 31.33 | 3.52 | .39** [.34, .43] | -- | | |
| 3. Causal terms (%) | 2.83 | 0.74 | -.07** [-.12, -.02] | .15** [.10, .19] | -- | |
| 4. Readability | 33.96 | 7.09 | .08** [.03, .13] | .02 [-.03, .07] | -.00 [-.05, .04] | -- |
| 5. Positive tone (%) | 0.58 | 0.34 | -.23** [-.28, -.19] | -.33** [-.37, -.28] | .02 [-.03, .07] | -.14** [-.19, -.09] |

*Note*. ** $p < .01$. Numbers in brackets are 95% Confidence Intervals.

**Table 2**

*Classification Model Results*

| Model | Features | Accuracy | Sensitivity | Specificity | AUC | Cohen's *h* |
|---|---|---|---|---|---|---|
| Obfuscation features | 5 | 60.6% | 66.9% | 54.4% | 0.627 | 0.214 |
| Full LIWC dictionary | 109 | 72.4% | 76.3% | 68.5% | 0.781 | 0.464 |
| Function word *n*-grams | 457 | 73.2% | 76.0% | 70.4% | 0.801 | 0.482 |
| Document embeddings | 768 | 73.9% | 78.4% | 69.5% | 0.811 | 0.499 |
| Open vocabulary approach | 9,087 | 75.6% | 77.4% | 73.9% | 0.830 | 0.538 |

*Note*. Accuracy results differ slightly here from the main text due to cross-validation. Function words (e.g., *n*-grams used by at least 5% of the training papers) were determined by *quanteda* (Benoit et al., 2018). Document embeddings were 768-dimension vectors from SPECTER (Cohan et al., 2020), a language model trained on scientific papers. Each paper was split into 200-word segments, and segment vectors were averaged to produce one vector per paper. This model used a ridge penalty ($\alpha = 0$) instead of an elastic net because embedding dimensions are dense and each carries some signal (Kjell et al., 2023). Open vocabulary *n*-grams were all unigrams and bigrams (i.e., two-word phrases) used by at least 5% of training papers. Cohen's *h* is the effect size for the difference between the model's accuracy and chance (50%).

**Figure 1**

*Obfuscation Results Over Time Across Language Dimensions*

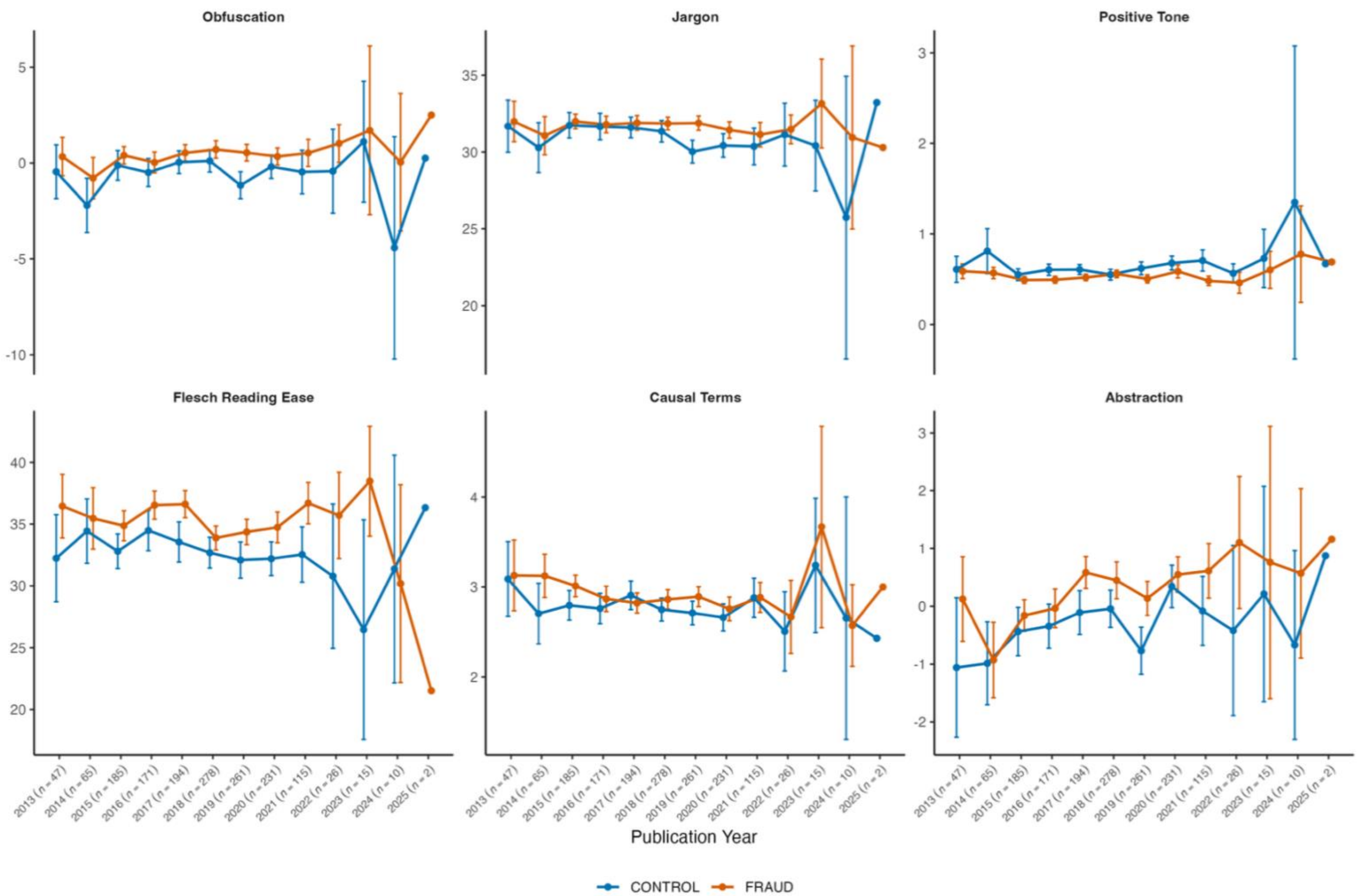

*Note*. Error bars are 95% Confidence Intervals. Sample size numbers combine fraudulent and genuine control cases.